%% file: main.tex
\documentclass[conference]{IEEEtran}
\IEEEoverridecommandlockouts
\usepackage{cite}
\usepackage{amsmath,amssymb,amsfonts}
\usepackage{algorithmic}
\usepackage{graphicx}
\usepackage{subfigure}

\usepackage{textcomp}
\usepackage{xcolor}
\def\BibTeX{{\rm B\kern-.05em{\sc i\kern-.025em b}\kern-.08em
    T\kern-.1667em\lower.7ex\hbox{E}\kern-.125emX}}

\usepackage{graphicx}
\usepackage[colorlinks=true, allcolors=blue]{hyperref}
\usepackage{listings}
\lstdefinestyle{Python}{
    language        = Python,
    frame           = lines, 
    basicstyle      = \footnotesize,
    keywordstyle    = \color{blue},
    stringstyle     = \color{green},
    commentstyle    = \color{red}\ttfamily
}
\usepackage{xcolor}

\definecolor{codegreen}{rgb}{0,0.6,0}
\definecolor{codegray}{rgb}{0.5,0.5,0.5}
\definecolor{codepurple}{rgb}{0.58,0,0.82}
\definecolor{backcolour}{rgb}{0.95,0.95,0.92}
\definecolor{circlecolor}{RGB}{196, 77, 116}

\lstdefinestyle{mystyle}{
    backgroundcolor=\color{backcolour},   
    commentstyle=\color{codegreen},
    keywordstyle=\color{magenta},
    numberstyle=\tiny\color{codegray},
    stringstyle=\color{codepurple},
    basicstyle=\ttfamily\footnotesize,
    breakatwhitespace=false,         
    breaklines=true,                 
    captionpos=t,                    
    keepspaces=true,                 
    numbers=left,                    
    numbersep=5pt,                  
    showspaces=false,                
    showstringspaces=false,
    showtabs=false,                  
    tabsize=2
}
\usepackage{tikz}
\newcommand*\circled[1]{\tikz[baseline=(char.base)]{%
            \node[shape=circle,fill=circlecolor!90,draw=white,inner sep=1pt] (char) {#1};}}
\usepackage{enumitem}

\title{Operator Fusion in XLA: Analysis and Evaluation}
\author{{Daniel Snider \qquad\quad Ruofan Liang} \\
University of Toronto
\thanks{This is a CSC2224 course project report.}\\
\{firstname.lastname\}@mail.utoronto.ca}
\date{}

\begin{document}
\maketitle

\begin{abstract}
Machine learning (ML) compilers are an active area of research because they offer the potential to automatically speedup tensor programs. Kernel fusion is often cited as an important optimization performed by ML compilers. However, there exists a knowledge gap about how XLA, the most common ML compiler, applies this nuanced optimization, what kind of speedup it can afford, and what low-level effects it has on hardware. Our paper aims to bridge this knowledge gap by studying key compiler passes of XLA's source code. Our evaluation on a reinforcement learning environment Cartpole shows how different fusion decisions in XLA are made in practice. Furthermore, we implement several XLA kernel fusion strategies that can achieve up to 10.56x speedup compared to our baseline implementation.
\end{abstract}

\input{intro}
\input{related_work}
\input{findings}

\input{evaluation}

\input{discussion}
\input{conclusion}

\bibliographystyle{plain}
\bibliography{main}

\end{document}

%% file: intro.tex
\section{Introduction}
Machine learning (ML) becomes more and more important in various computer tasks including computer vision, natural language processing, robotic control, etc. The computation efficiency of ML also arises as a popular topic for computer system and architecture research.
Today’s machine learning (ML) applications rely on specialized hardware accelerators like GPUs in order to be performant. 
However, because hardware accelerators are designed to be simple and fast, they lack features that CPUs have to automatically speed up user code. As a result of this and with the addition of the complexity of designing efficient parallel programs, modern ML programs require complex optimizations. Furthermore, these optimizations are often unique to each combination of algorithm, data shape, and hardware.  To address this problem, ML compilers can make these optimizations automatically, but they are still limited and an emerging field of research. The ML compilers take the model definitions described in the ML frameworks as
inputs, and generate efficient code implementations on various ML hardware as outputs. The transformation between model definition and specific code implementation are highly optimized targeting the model specification and hardware architecture \cite{li2020deep}.

The adoption of ML compilers for GPUs is still nascent. 
The existing works like TVM AutoScheduler \cite{chen2018tvm}, Ansor \cite{zheng2020ansor}, Rammer \cite{ma2020rammer}, and TorchScript compiler \cite{TorchScript} are limited to supporting ML inference. On the other hand, XLA compiler \cite{XLA} is more mature and flexible. 
XLA is the most widely used ML compiler because it is used in TensorFlow \cite{abadi2016tensorflow} and JAX frameworks \cite{frostig2018compiling}. 
A common optimization step for XLA and other ML compilers is operator fusion. ML compilers can automatically fuse multiple operations into one computation kernel to reduce the memory transfer and kernel launch overhead. XLA uses rule-based fusion strategies to fuse operations that meet certain fusion patterns and requirements set by experienced developers.
Recent research often compares XLA's performance to hand-optimized GPU kernels like cuDNN and super-optimizations like Ivanov et al. \cite{ivanov2021data}. These papers find performance gaps between their optimized kernels and XLA's generated kernels. 
We are motivated to look closely at this performance gap by investigating what optimizations and sacrifices XLA makes in its approach to automatic optimization.

By analyzing XLA's source code and evaluating a microbenchmark of XLA, we are now able to give an in-depth explanation on XLA's operation fusion behavior. XLA's fusion heavily depends on the initial computation graph converted from the Python frontend. A low-quality python implementation can cause a major performance drop. Additionally, conservative fusion criteria in XLA also limits the opportunities for optimization. Our evaluation on a reinforcement learning environment Cartpole \cite{sutton2018reinforcement} shows how different fusion decisions in XLA are made in practice. Furthermore, we also explore some potential optimizations that can achieve up to 10.56x speedup compared to our baseline JAX-XLA implementation.

To summarize, we make the following contributions in this project:
\begin{itemize}
    \item We give an in-depth introduction to the fusion mechanism of XLA, which has not yet been well described in public papers or documents.
    \item We evaluate XLA's performance with one JAX-based RL task and explore ways to further improve the performance of XLA's generated code.
    \item We make a low-level profiling of XLA programs to identify the existing limitations and potential optimizations for the XLA compiler.
\end{itemize}

%% file: related_work.tex
\section{Related Work}
% \subsection{Kernel Fusion}
\subsection{Optimized ML Frameworks}
The conventional ML frameworks such as PyTorch \cite{paszke2019pytorch}, TensorFlow (without XLA) \cite{abadi2016tensorflow}, and MXNet \cite{chen2015mxnet} commonly map DL operations to cuDNN/cuBLAS primitives or pre-implemented CUDA kernels for the full flexibility of ML algorithm prototyping, but this design cannot provide full optimizations for ML programs.
The optimized deep learning frameworks are later proposed to generate or use workload-specific kernels for better performance. The typical DL optimization frameworks include XLA \cite{XLA}, TensorRT \cite{TensorRT}, TVM \cite{chen2018tvm}, Tensor Comprehensions \cite{vasilache2018tensor}, etc.  XLA and TensorRT use some manually defined rules to fuse simple operations, while for complicated operators such as convolution, matrix multiplication, these frameworks still rely on the cuDNN/cuBLAS primitives. One advantage of this type of ML compilers is that they require relatively shorter compile time to finish optimization.
On the other hand, frameworks like TVM and Tensor Comprehension have more flexible code generation, instead of using primitive kernels, these frameworks can automatically tune the fused kernels by using some learning algorithm such as GBM, simulated annealing, and genetic algorithm, which can result in relatively longer compilation time and limited sub-graph level searching and tuning.

Compared to other ML compilers, XLA is one of the most widely-used ML optimization compilers because of its core position in Google's AI computation environment from the frontend (TensorFlow/JAX) to the backend (TPU). The developers can directly benefit from the speedup provided by XLA from their TensorFlow or JAX code without additional effort on the code conversion. This is also one the main reasons we choose XLA as our target ML optimization framework.

% \subsection{XLA Frontend}
% JAX \cite{frostig2018compiling} is a domain-specific tracing JIT compiler for generating high-performance accelerator code from pure Python and Numpy machine learning programs. JAX directly uses the XLA compiler to generate optimized code for acceleration on GPU/TPU. With the help of Autograd, JAX can do automatic differentiation for DNN model training.

\subsection{Operation Fusion in DL}
The operation fusion is also an active topic for ML compiler research. In addition to aforementioned ML optimization frameworks, recent works explore more complicated fusion strategies for better hardware resource utilization.

Ivanov et~al. \cite{ivanov2021data} do an exhaustive search of all possible data layout and operator fusion in the Transformer model \cite{vaswani2017attention}. Their results show that an optimal fusion strategy can provide up to 22.91\% data movement reduction and overall achieve a 1.30x performance improvement over the state-of-the-art implementations. 
PET \cite{wang2021pet} allows partially equivalent transformations such as joining two tensors and performing a single convolution instead of two. Such equivalent transformation can increase the optimization search space, and PET uses the beam search to optimize the entire DNN.
DeepCuts \cite{jung2021deepcuts} applies a greedy exploration guided by an analytical cost model to tune graph-level fusion decisions along with some CUDA kernel parameters. However, DeepCuts only considers limited vertical operation fusions in the computation graph, and it has relatively high implementation cost for the new operations because of its analytical cost model.
Unlike the XLA compiler, \cite{ivanov2021data,wang2021pet}, and \cite{jung2021deepcuts} all employ some types of design space search algorithms to find some better fusion/optimization combinations. These methods require much longer compilation time (it can be multiple hours) to find a good enough solution, thus algorithm developers cannot instantly benefit from these optimizations during the fast algorithm prototyping stage. The equivalent computation transformation mentioned in \cite{wang2021pet} and \cite{niu2021dnnfusion} could be a promising future optimization direction, which will also be discussed in our XLA analysis.

There are other non-trivial recent ML fusion optimization works. For example, XTAT \cite{phothilimthana2021flexible} uses the auto-tuning method directly on the XLA's multi-pass optimization to generate better kernels for TPU. XTAT can perform joint optimization for different XLA passes at both graph-level and subgraph-level. However, this autotuner is relatively compute-heavy and time-consuming, which cannot be finished by the JIT compilation of TensorFlow/JAX. Besides, XTAT autotuner only targets TPU accelerators, and is mainly used for improving the existing heuristic algorithm in Google's internal TPU-XLA. 

HFTA \cite{wang2021horizontally} is another fusion technique that vectorizes the training of multiple models together. This reduces overhead (number of CUDA API calls, DL framework GPU memory footprint, etc.) for GPU computation which in turn increases throughput. Though XLA also has some horizontal fusion mechanisms, HFTA supports end-to-end training and is an ideal way to accelerate hyperparameter tuning.

%% file: findings.tex
\begin{figure*}[h!]
	\centering
	\subfigure[Instruction Fusion.\label{fig:inst_fusion} ]
	{\includegraphics[width=0.14\textwidth]{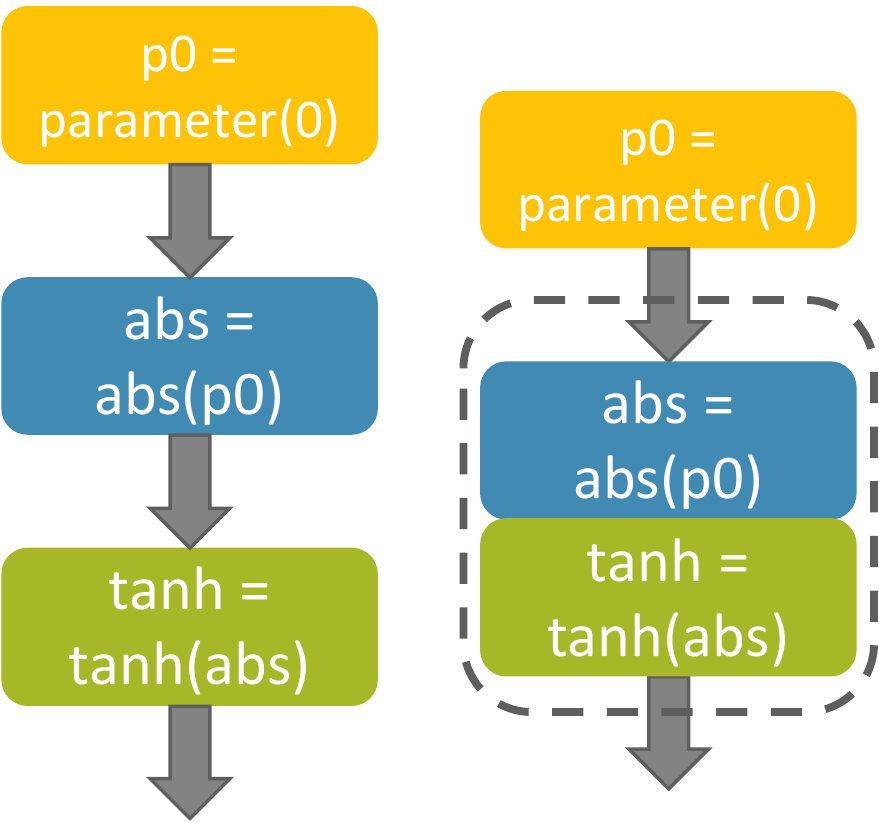}}
	\subfigure[Fusion Merger. \label{fig:fusion_merger} ]
	{\includegraphics[width=0.28\textwidth]{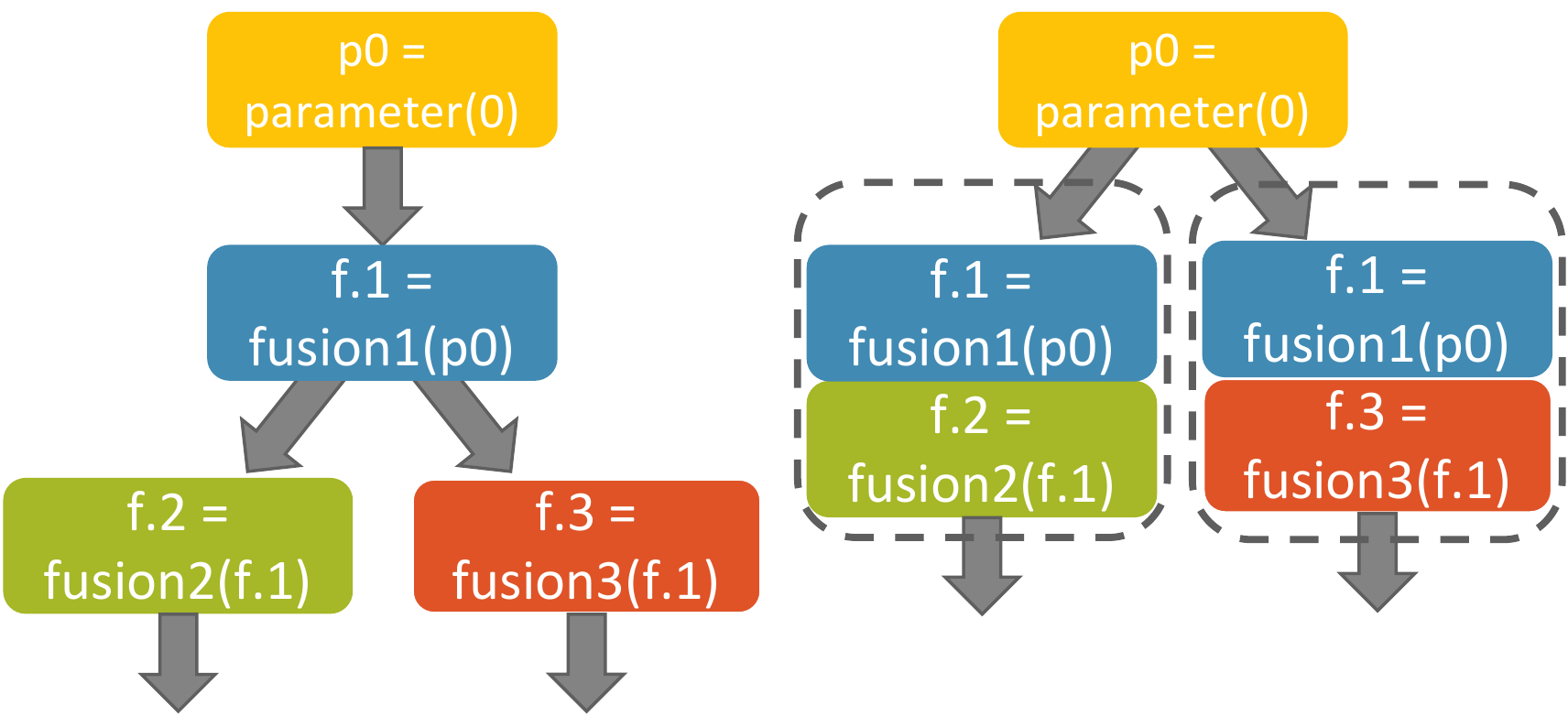}}
	\subfigure[Sibling Fusion. \label{fig:sibling} ]
	{\includegraphics[width=0.28\textwidth]{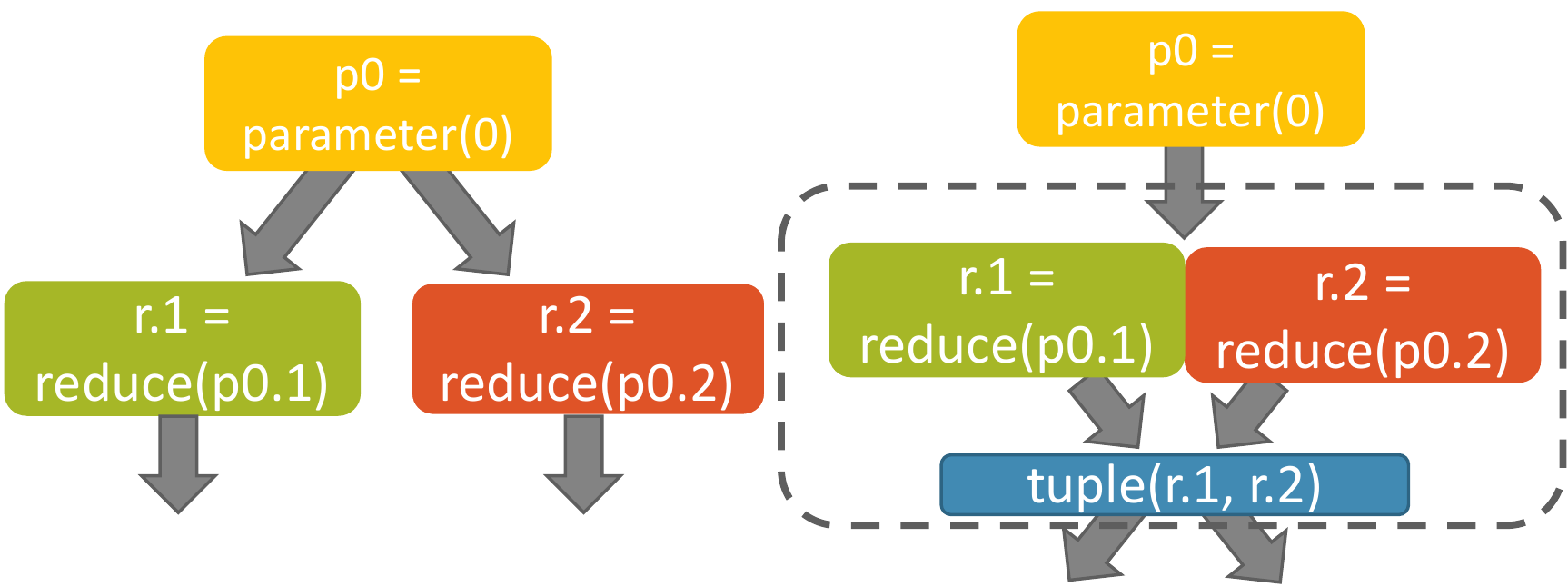}}
	\subfigure[Producer-consumer Fusion. \label{fig:producer} ]
	{\includegraphics[width=0.25\textwidth]{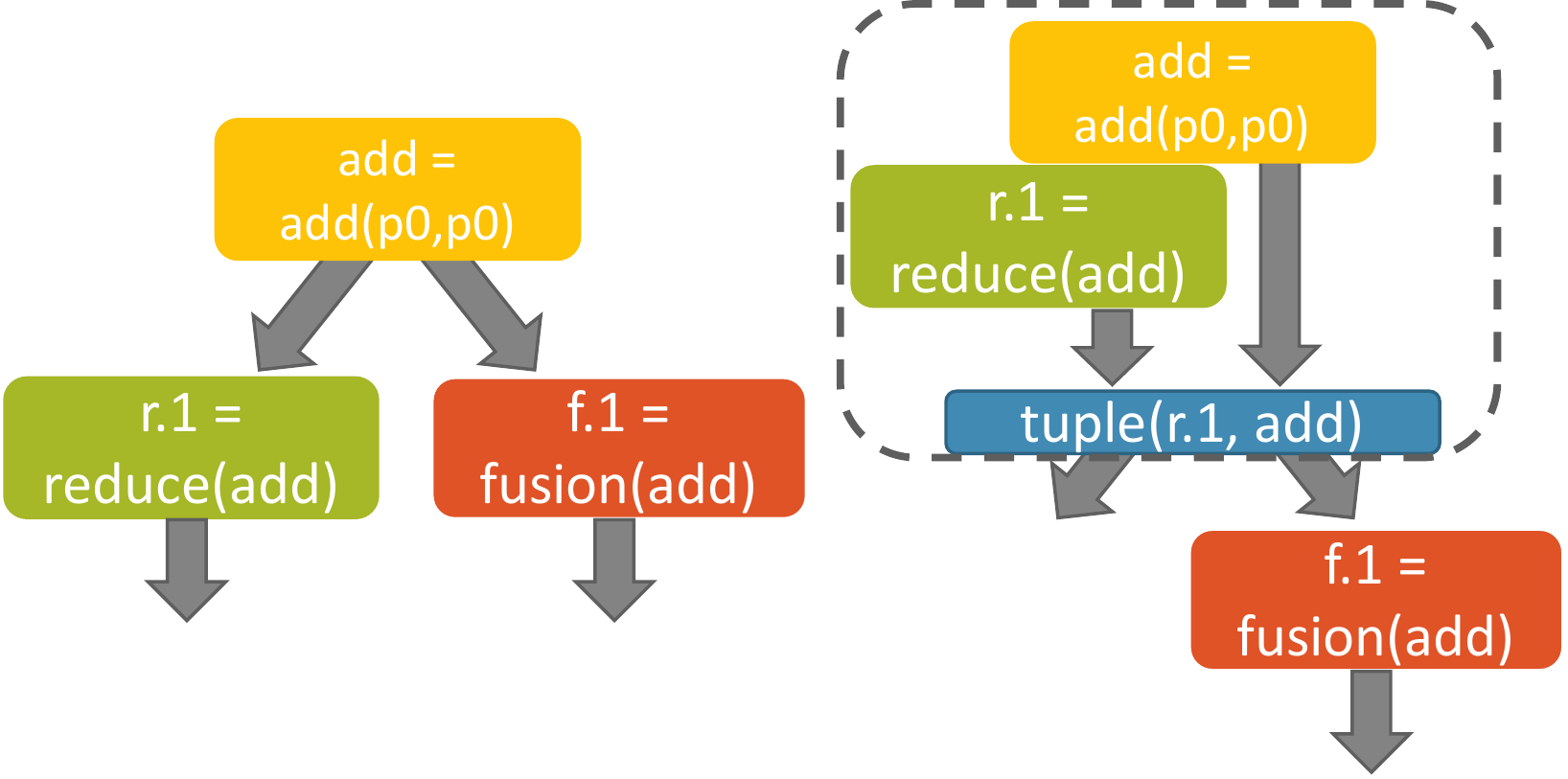}}
	\vspace{-7pt}
	\caption{Fusion strategies in XLA. \label{fig:xla_fusion}}
\end{figure*}

\section{XLA's Multi-pass Optimization}

% \paragraph{JAX Design Goals} It should be noted that JAX \cite{frostig2018compiling} is designed to be flexible, performant (using XLA), and have a low compile-time overhead on the order of seconds.

\subsection{XLA Computation Graph Optimization}\label{xla_passes}
Once the traced computational graph from TensorFlow or JAX is sent to the XLA compiler, a series of fine-grained optimization passes are executed to gradually optimize the initial computation graph (represented as an XLA HLO IR). Some optimization passes are performed more than once, with the most common being Dead Code Elimination (DCE) and Common Subexpression Elimination (CSE). Optimization passes are logically grouped into what are called "\texttt{Pass Pipelines}". XLA's graph-level optimization pipeline passes are listed below. Our experimentation has found that kernel fusion is one of the last optimization pipelines to run.
\begin{itemize}
    \item \textbf{SPMD partitioner}. This step partitions tensors to be operated on in parallel across devices. (SPMD, Single Program Multiple Data) \cite{lepikhin2020gshard}.
    \item \textbf{Optimization}. Includes passes for canonicalization, expansion, and simplification. Mostly simple rule-based operator conversions for later use, e.g., BatchNorm Expander and Logistic Expander converts convert a complicated computation into a sequence of simple operations).
    \item \textbf{Simplification}. Performs simplification to specific operations, e.g., inlining and constant propagation, and WhileLoopSimplifier removes dead tuple elements.
    \item \textbf{Collective optimizations}. Optimizes collective operations (e.g., reduce, gather) generated by SPMD partitioning for multiple devices.
    \item \textbf{Conv canonicalization}. Canonicalize convolution operations, i.e., reshaping the input and filter to NHWC and HWIO order, respectively.
    \item \textbf{Layout assignment}. Pre-assigns layouts of some operands to satisfy layout constraints and results of library calls (e.g., cuDNN, cuBLAS).
    % \item \textbf{nvptx post-layout_assignment}
    \item \textbf{Post layout assignment} Performs target-specific HLO optimization passes after layout assignment, e.g., Optimize padding for cuBLAS and pick GEMM or Conv algorithm.
    \item \textbf{Fusion} Performs different types of vertical operation fusion, e.g., simple instruction fusion, fusion merger, and multi-output fusion.
    \item \textbf{Horizontal fusion}. Performs the horizontal operation fusion. It includes horizontal loop fusion and horizontal input fusion.
    \item \textbf{Post fusion optimization}. Combines small non-dependent collective operations into larger combined operations.
    \item \textbf{GPU IR emit prepare}. Sanitizes the given HLO module so that it will be accepted by IR Emitter.
\end{itemize}

In this project, we mainly focus on the fusion  optimization passes that help XLA gain speedups. Fusion strategies are discussed in detail in the next subsection.

\subsection{Operation Fusion}
Because XLA's fusion strategies are not described in any official documentation or paper we have reviewed the XLA source code \footnote{\url{https://github.com/tensorflow/tensorflow/tree/master/tensorflow/compiler/xla}} in-depth. We have extracted all the fusion strategies used by XLA.
Fig. \ref{fig:xla_fusion} shows four typical fusion strategies commonly used by XLA.
More details about different types of kernel fusions are listed as follows.

\textbf{Instruction Fusion}. This is a simple vertical operation fusion step, in which producing instructions are fused into their consumers with the intent that the sequential operations will be fused in code generation (see Fig. \ref{fig:inst_fusion}). XLA does a reverse post-order traversal in this step to determine whether two dependent operations should be fused or not via a \texttt{ShouldFuse} function. XLA defines several rules to check whether the operation is fusible. For example, XLA explicitly maintains a list of "expensive" operations\footnote{\url{https://github.com/tensorflow/tensorflow/blob/master/tensorflow/compiler/xla/service/instruction_fusion.cc##L62}} (e.g., convolution, sort, all reduce, etc.) that should not be fused. XLA also checks whether the fused kernel will be too large for the  GPU, whether the fused kernel will cause a nested loop, etc. XLA will make sure not to exceed several GPU hardware limits including threads per block, shared memory per block, and threads per SM.

\textbf{Fusion Merger}.
This fusion pass attempts to merge fusion instructions to reduce memory bandwidth requirements and kernel launch overhead (Fig. \ref{fig:fusion_merger}). Fusion instructions are merged into their users if some conditions are met. For example, the result of merging the fusion instruction into its users would not increase bytes transferred; and if producer operations are fusible with all consumers (if they are not fusible with at least one consumer, they won't be fused at all).

\textbf{Multi-Output Fusion}. Multi-output fusion of sibling and producer-consumer instructions for the GPU backend is also intended to reduce memory bandwidth requirements. Typically, there are two types of multi-output fusion this pass performs--sibling multi-output fusion (Fig. \ref{fig:sibling}) and producer-consumer multi-output fusion (Fig. \ref{fig:producer}). 
Fusion of sibling operations can reduce memory bandwidth requirements, because common input parameters have to be read only once.
Fusion of producer-consumer operations reduces memory bandwidth requirements by eliminating one read from memory. %In the example above, B does not need to read the output of A from memory, while C still does (using gte\_a).
Sibling fusion and producer-consumer fusion can usually meet the fusion constraints at the same time. XLA will select the one that can give more fusion opportunities for later fusion optimizations, and sibling has a higher priority over producer-consumer by default.
These two types of multi-output fusion can also be combined in this fusion pass.

\textbf{Horizontal Fusion}. This fusion pass horizontally fuses computations to reduce kernel launch overhead while increasing kernel launch dimensions on the GPU. 
The initial motivation of the horizontal fusion is due to the observation that the training optimizer phase (e.g., Adam optimizer and L2Loss, etc.) typically has many small kernels as a result of applying the same formula on many training parameters (or variables in Tensorflow). Fusing these small kernels, hence, provides performance gain.
This fusion style provides an important advantage that kernels of different shapes can be horizontally fused\footnote{See here for more details \url{https://github.com/tensorflow/tensorflow/blob/master/tensorflow/compiler/xla/service/gpu/horizontal_loop_fusion.h##L63}}. For example, consider a multiply operation and an add operation with separate input shapes, but their output is consumed by a common operation, here horizontal fusion is triggered.

\subsection{Other Findings}

\textbf{Kernel scheduling and CUDA streams}. At compile time, XLA's \texttt{IrEmitter} also generates \texttt{KernelThunks} which contain necessary arguments for launching kernels. At runtime, \texttt{GpuExecutable} launches the kernel using the \texttt{KernelThunk} which specifies the buffer addresses of the data needed for the kernel launch. An initial finding is that the function \texttt{BFSLaunchOrder()} computes a topological launch order that is close to a breadth-first order. This enables the possibility of launching kernels concurrently in different CUDA streams. The function \texttt{CanRunConcurrently()} returns whether the two HLOs can run concurrently, however in practice we have not seen multiple streams utilized by XLA.

\textbf{JAX Compile-time speed}. XLA's compile time can be slow, especially when the input JAX source code includes native python loops. Each python loop iteration is traced and adds a repetition of body into the XLA intermediate representation. To avoid this a user must use JAX's built-in loop constructs. Through detailed investigation of Nsight System execution timelines we have identified two ways to speed up JAX's compile time that are not reported anywhere that we have seen: (1) disabling JAX tree flatten conversion and operate only on arrays (not objects), or (2) disable GEMM autotuning because it often doesn't help.

\textbf{Rules-based optimization}. XLA compiler does not search for any optimization, and the number of compiler passes is fixed. The trade-off between time spent waiting for compilation and compiler-based speedups has been chosen to work well for general purpose, everyday development. This trade off can be controlled, but the interface is not easy to control for normal uses. You can disable compiler passes using an environment variable, but this requires you to know the low-level names of XLA's compiler passes.

%% file: evaluation.tex
\section{Case Study on Cart-pole}

\begin{figure}
    \centering
    \includegraphics[width=\linewidth]{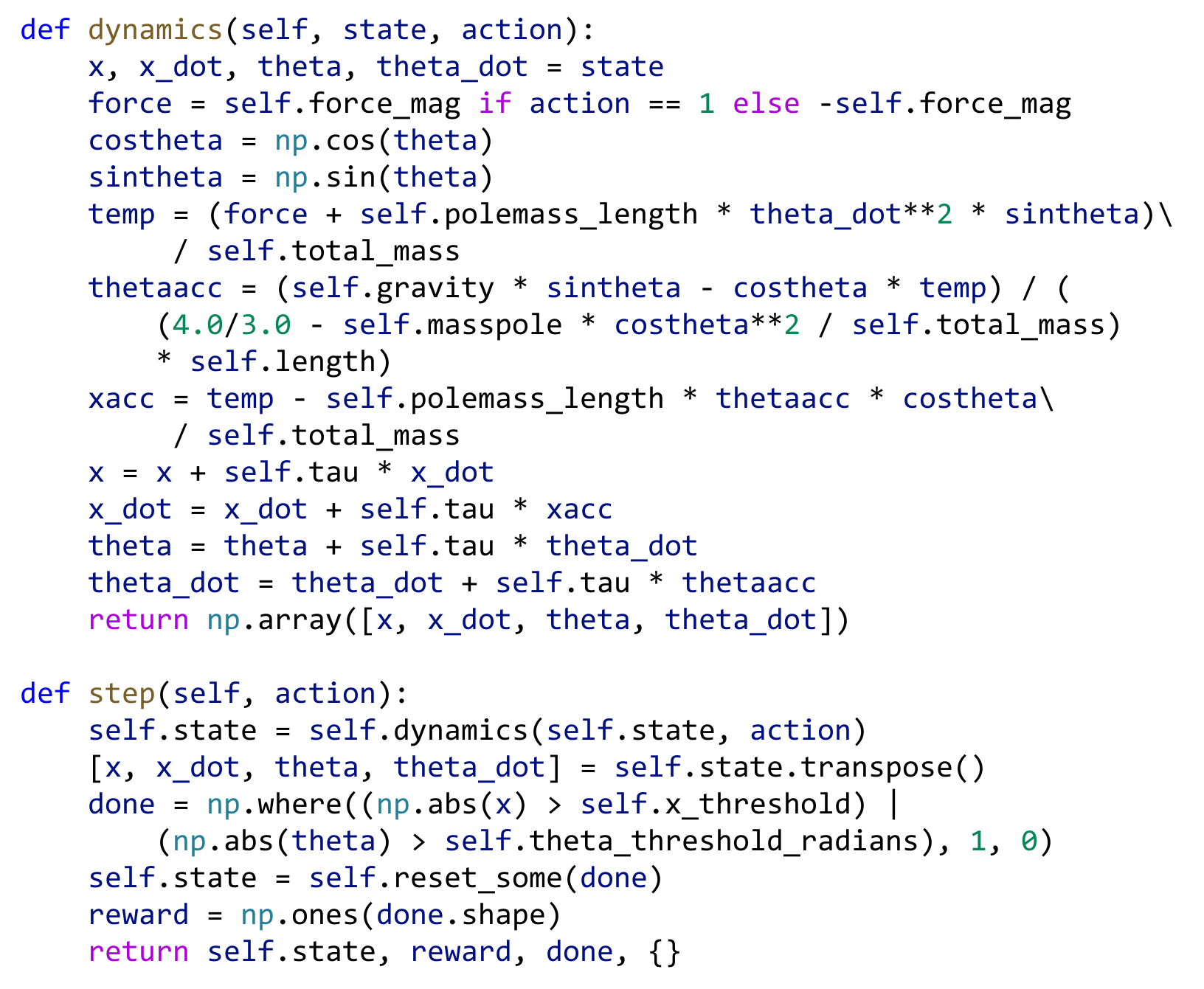}
    \caption{The JAX code for the Cart-pole environment update step.}
    \label{fig:cartpole_py}
    \vspace{-8pt}
\end{figure}

We use a classic reinforcement learning environment Cart-pole \cite{sutton2018reinforcement} as the benchmark task to study the XLA graph optimizations (this is inspired by the recent success of JAX-based RL simulation framework \cite{brax2021github}). 
Cart-pole is a dynamic control system with an inverted pendulum on a cart.
We implement the Cart-pole environment in JAX, Fig. \ref{fig:cartpole_py} shows our baseline implementation of the environment update function, the core computation of our benchmark task. We will show how XLA converts this code into fused kernels as well as further analysis and optimization of this code in the following sections.

\subsection{XLA Compilation}
\begin{figure*}
	\centering
	\subfigure[Before any optimization.\label{fig:before_optimization} ]
	{\includegraphics[height=20em]{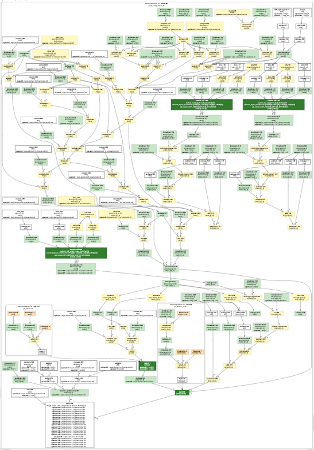}}
	\subfigure[Before fusion optimization\label{fig:before_fusion} ]
	{\includegraphics[height=20em]{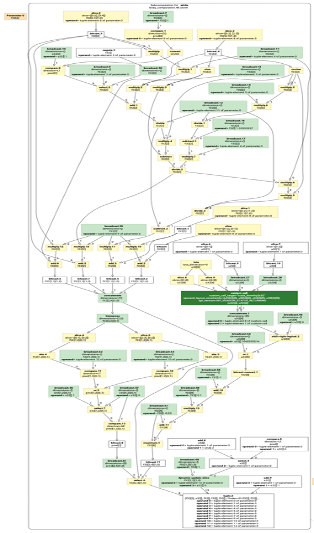}}
	\subfigure[After the fusion optimizations; right: 3 fusion boundaries for the fused kernels.\label{fig:after_fusion} ]
	{\includegraphics[width=0.45\linewidth]{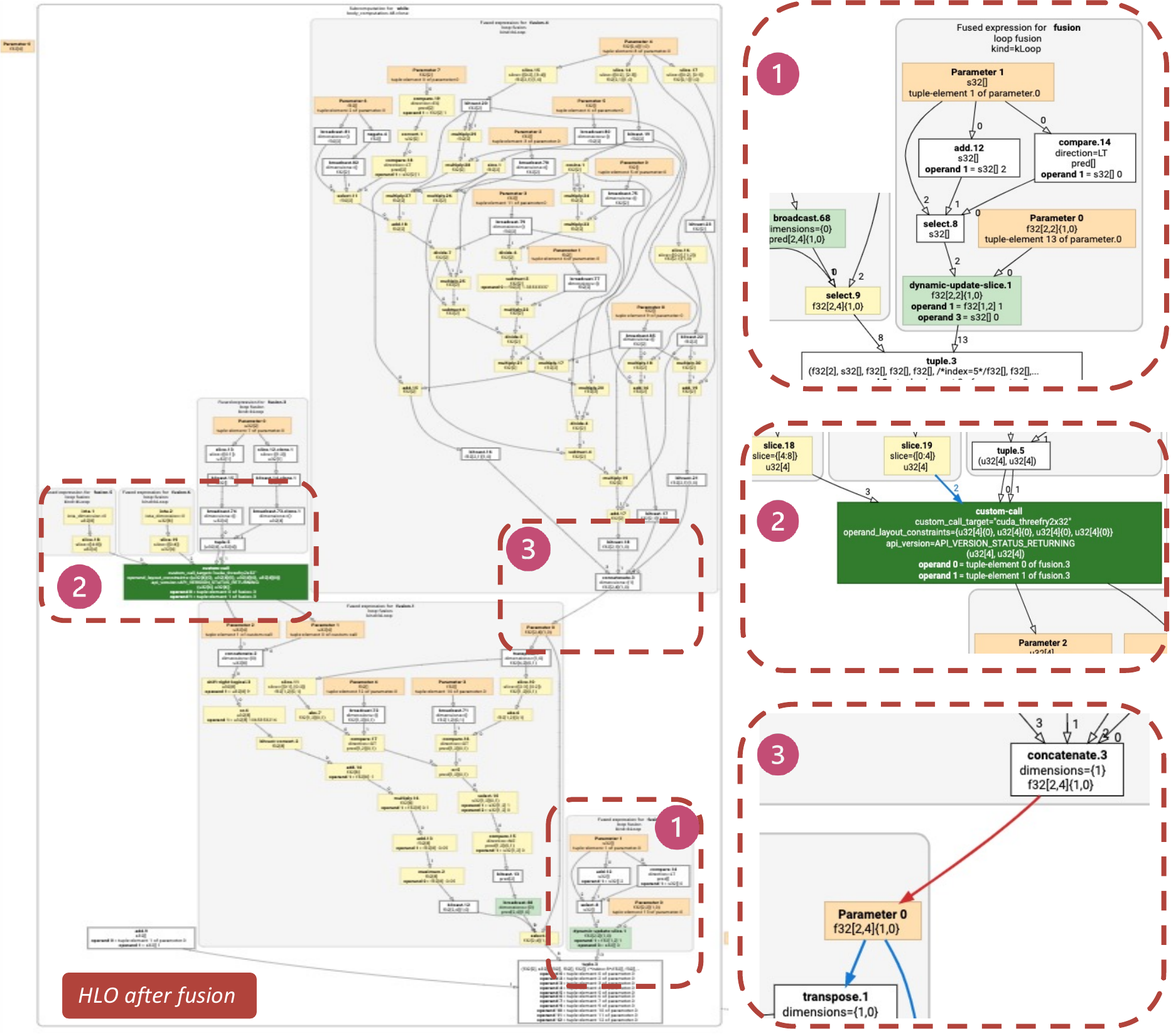}}
	\caption{HLO computation graph for Cartpole update step.\label{fig:hlo_cartpole}}
	\vspace{-7pt}
\end{figure*}

When running our JAX-Cartpole simulation environment, the JAX JIT compilation first traces and converts the Python code into the HLO computation graph. This HLO IR then goes through multiple optimization passes, as described in Sec. \ref{xla_passes}. Fig. \ref{fig:before_optimization} shows the initial HLO graph converted by JAX, and Fig. \ref{fig:before_fusion} shows the HLO graph after several optimization passes (e.g., simplification, layout assignment optimization, etc.) and right before the fusion optimization pass.
By comparing the two computation graphs, we can see that XLA removes or replaces the duplicate or redundant operations from the initial HLO graph in a multi-pass manner. Because our focus is on the operation fusion optimization, we skip the detailed discussion of XLA optimization passes before the fusion in this report. Next, we will explain XLA's fusion decisions performed on the HLO graph \ref{fig:before_fusion}.

% \subsection{Fusion Optimizations}
% \begin{figure}
%     \centering
%     \includegraphics[width=0.9\linewidth]{figs/hlo_after_fusion.pdf}
%     \caption{Left: HLO graph after the fusion optimizations; right: 3 fusion boundaries for the fused kernels.}
%     \label{fig:after_fusion}
% \end{figure}

Fig. \ref{fig:after_fusion} shows the HLO graph after all the fusion passes. We see from the graph that operations in graph \ref{fig:before_fusion} are fused into 6 fused kernels. To understand how XLA makes the current fusion decisions, we investigate the XLA's source code and runtime log file. 
We find that most of operations in the graph are fused in the instruction fusion pass, and there three interesting fusion boundaries are shown in the graph (the boxes with dashed lines in \ref{fig:after_fusion}) which form final set of fused kernels. Next we  explain these three XLA fusion boundaries.
\begin{enumerate}[label=\protect\circled{\color{white}\arabic*}]
    \item The first one is at the bottom of the computation graph. The fused kernel shown in box 1 is used only for XLA's implementation of while loop control. By looking at XLA's fusion code\footnote{ \url{https://github.com/tensorflow/tensorflow/blob/master/tensorflow/compiler/xla/service/gpu/gpu_fusible.cc##L215}} we find that XLA does not fuse a tuple into its producer because it does not provide performance gain. This is because a tuple is not a kernel operation, it is a location in global memory. This tuple is used as the output of each loop step and the final output when the loop terminates. 
    \item The second one is in the middle left of the computation graph. This fusion boundary involves a custom-call operation "\texttt{cuda\_threefry2x32}", which is a pre-defined kernel for the random number generation. XLA does not have the ability to fuse such custom operations into its consumer or producer. Other common custom-calls such as cuDNN/cuBLAS primitives can also halt the expansion of fused kernels at these operations.
    \item The third one is in the middle right of the computation graph for the concatenation operation. Fusing \texttt{concatenate} in this case violates the XLA's predefined fusion rules that a \texttt{concatenate} operation with more than one user (blue arrows in the graph) cannot be fused, because XLA developers think such fusion may cause potentially high code duplication in general cases. However, we do not think the \texttt{concatenate} operation in our HLO graph could cause relatively high code duplication. We will show our attempt to bypass this limitation by modifying XLA's fusion decision functions.
\end{enumerate}

\section{Evaluations}

We performed 6 experiments to gather insights about XLA's fusion behaviour and performance. In each experiment we measured 2048 parallel simulation environments for 10,000 steps. We chose 2048 because it is the largest number of environments used in the Brax paper to make effective use of the GPU, while not being so large to cause RL learning to diverge. We choose 10,000 steps because it is enough to amortize the one-time overhead caused by the JAX framework which includes serializing objects and moving data to the GPU to begin execution. We tested on the Eco-13 server which contains an RTX 2080Ti GPU with CUDA driver 11.2.

\begin{figure}[!b]
\centering
  \includegraphics[width=1\columnwidth]{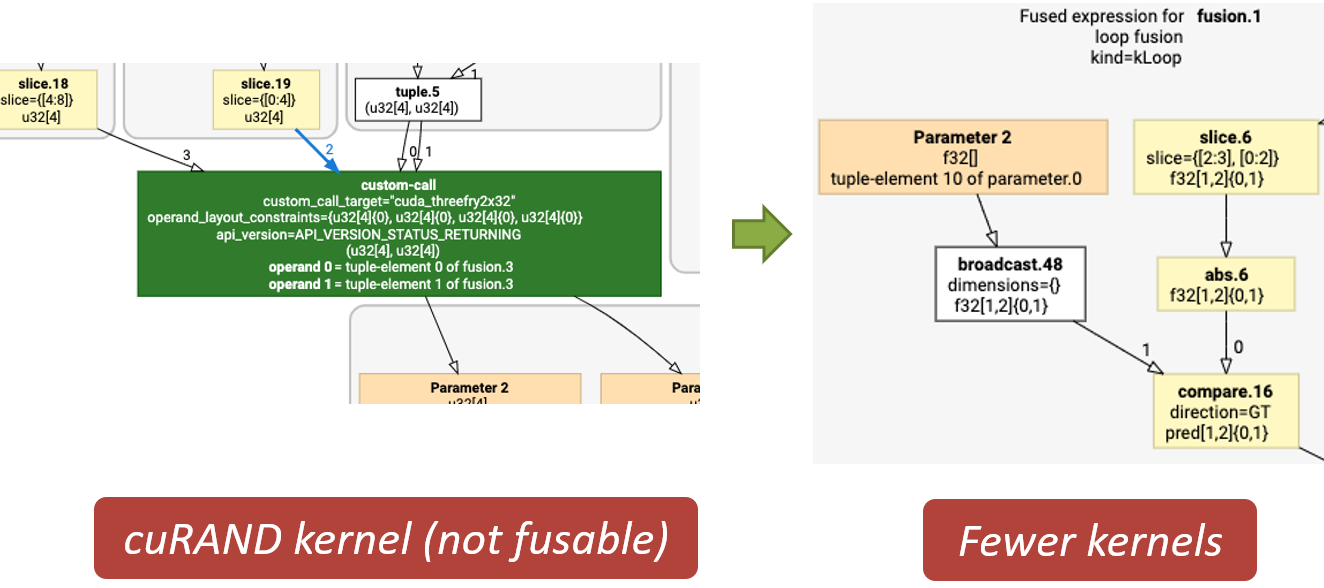}
  \caption{We replaced the cuRAND kernel (in green) with precomputed random values to bring our cartpole implementation closer to a single fully fused kernel. }~\label{fig:remove_random}
\end{figure}

\subsection{Remove cuRAND Kernels (Baseline)}

\begin{figure*}[htbp]
    \centering
	\includegraphics[width=0.8\textwidth]{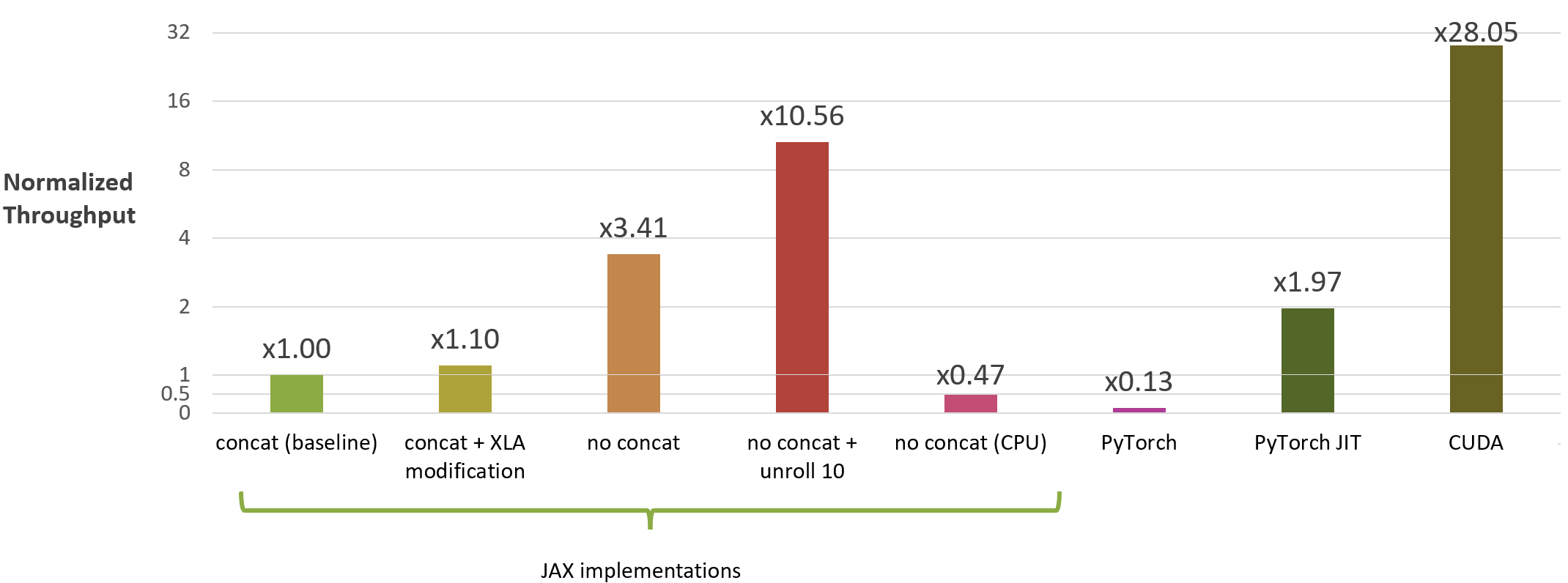}
	\vspace{-5pt}
	\caption{Normalized throughput of different implementations of cartpole simulation.\label{fig:all_eval}}
\end{figure*}

The first thing we did was to remove the unfusable "cuda\_threefry" cuRAND kernel which is responsible for randomness. We did this by precomputing a pool of random values to be used as random actions in the simulator and random start states for environment resets. The result was a reduction of the cuRAND kernel and it's 3 parent kernels as seen in figure \ref{fig:remove_random}. This yielded a 1.87x speedup. What remains are four kernels. Two kernels for the simulation and two kernels involved in JAX's implementation of the "scan" loop. In our throughput plot, Figure \ref{fig:all_eval}, this implementation corresponds to "concat (baseline)".

\subsection{Fusion via XLA Modification}
\begin{figure}[h]
\centering
  \includegraphics[width=.85\columnwidth]{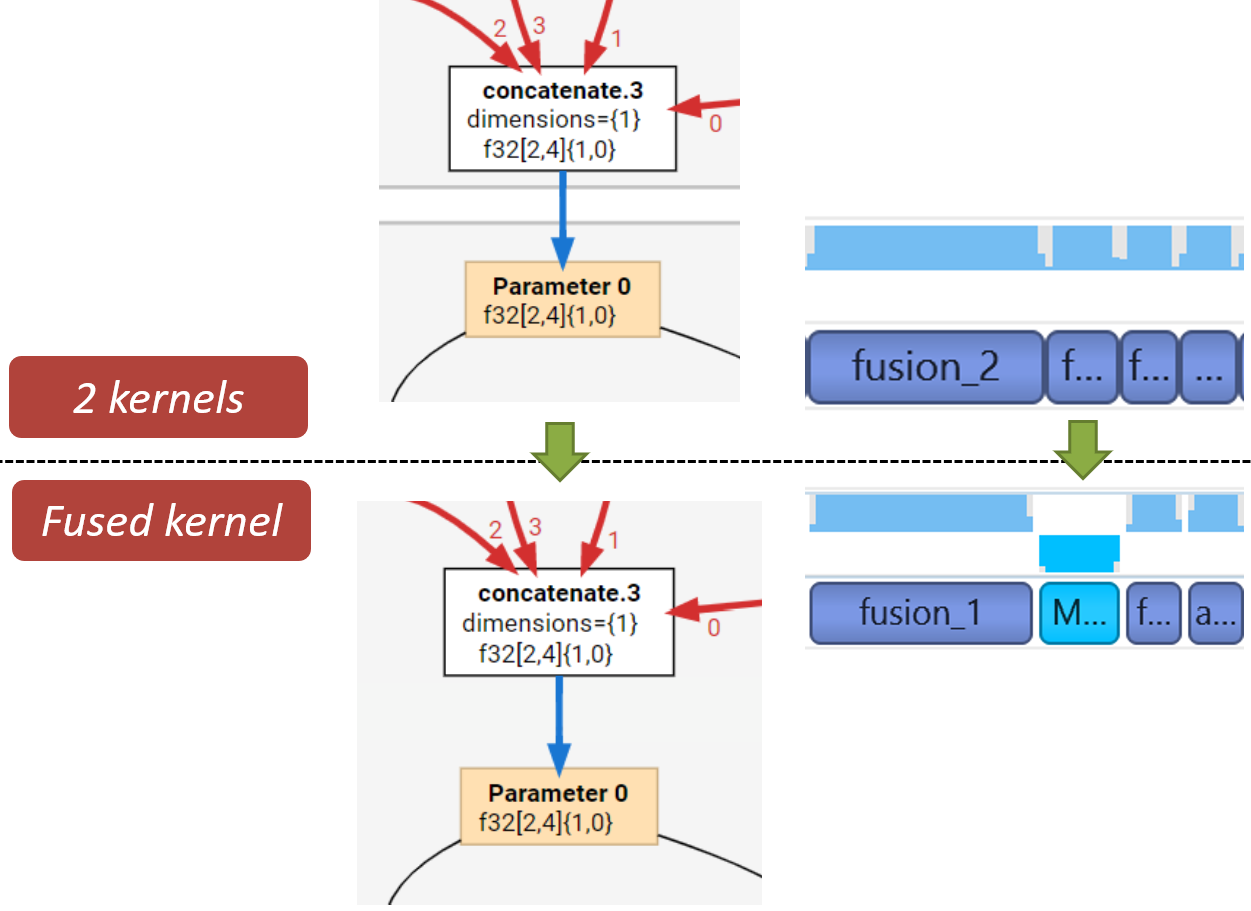}
  \caption{We modified XLA so that it would fuse this concatenation operation into its child operation.}~\label{fig:xla_mod}
  \vspace{-6pt}
\end{figure}

With two kernels left to run the simulator, we investigated why XLA didn't fuse them. We found that the situation resembled the Fusion Merger type seen in Figure 1. XLA's prevented this fusion because of a function called function CodeDuplicationTooHigh(). Fusion merger requires the parent operation to be fused with the child, however, there are two consumer operations, each needing the parent to be duplicated, and it appears that XLA's code doesn't want to support this. We modified XLA's source code to allow up to three consumers of the parent operation which led to fusion of the two simulation kernels (see Figure \ref{fig:xla_mod}). 

However, we only saw a marginal 10\% speedup. It is important to note that the operation that was separating the two kernels was a "concatenate" operation which writes a new array that is too large to fit into registers. We hypothesize that memory movement is a larger bottleneck than launch overhead, and because we haven't changed the amount of memory operations, the speedup is negligible. Using Nsight Systems we confirmed that one kernel was eliminated by in its place additional Device-to-Device memory transfer remained.

\subsection{Fusion via Memory Movement Optimization (no concat)}
\begin{figure}[h]
\centering
  \includegraphics[width=1\columnwidth]{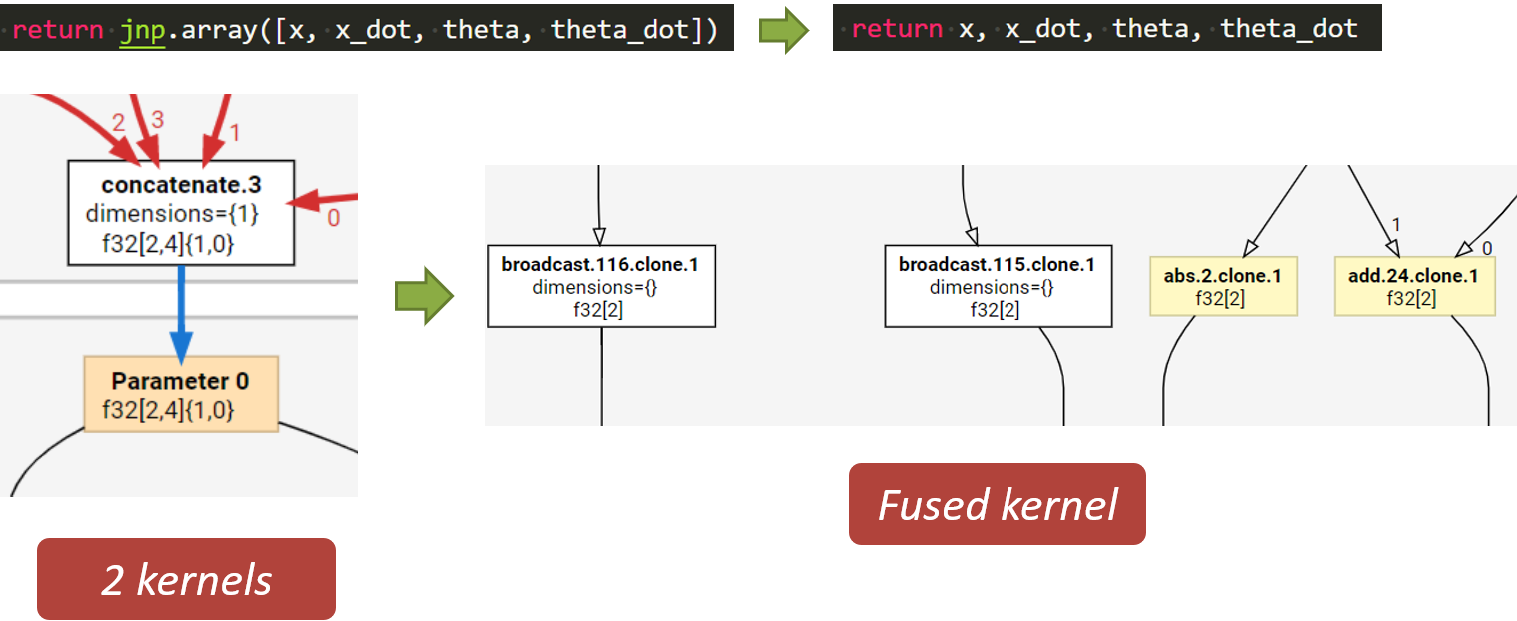}
  \vspace{-10pt}
  \caption{We improved our code to remove the concatenation operation which allowed XLA to fuse together two cartpole kernels into one. }~\label{fig:no_concat}
\end{figure}

Next we approached kernel fusion by addressing the core problem of our simulator's design. The concatenation operation was an unnecessary convenience to communicate state between our functions, so we instead passed the four state values individually as seen in Figure \ref{fig:no_concat}. 

This memory movement optimization allowed XLA to fully fuse the simulation and the variables could remain local in registers without needing to be combined at a higher level of memory. This resulted in a 3.41x speedup. Using Nsight Compute we confirmed that register usage went up by 40\%, the total executed instructions went down by 50\% (most of them memory requests), and the number of stalled cycles went down by 33\%. 

\subsection{Fusion via Unroll}
At this point our simulation kernel was still characterized as a tiny kernel full of elementwise operations which made it bound by kernel launch overhead. Our next strategy to kernel fusion was to use the built-in support to unroll Jax's scan loop. We unrolled the loop 10 times which means that one kernel contains the duplicated instructions for 10 loop iterations. 

\begin{figure}[h]
\centering
  \includegraphics[width=1\columnwidth]{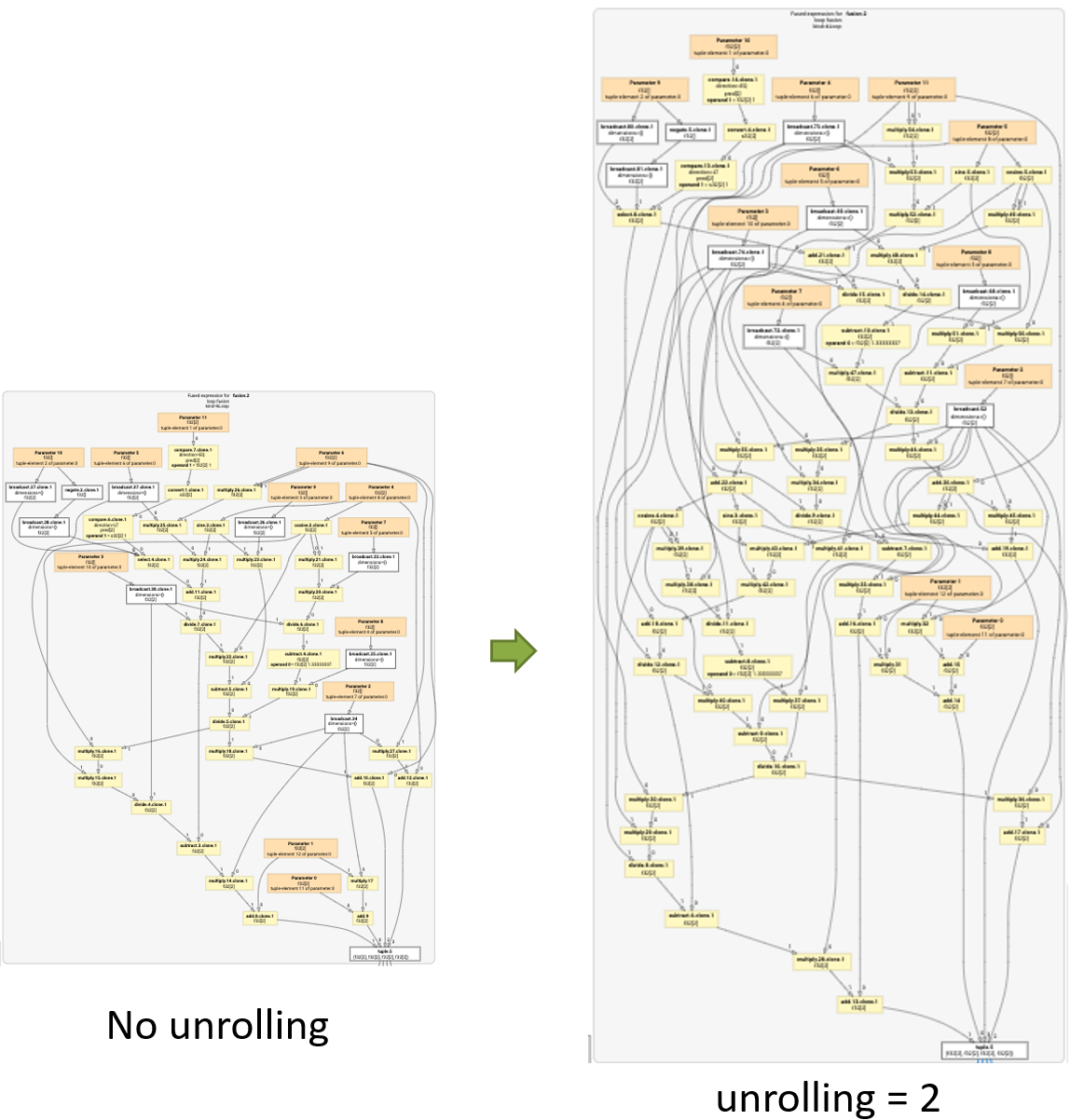}
  \caption{Illustration of XLA's HLO IR computational graph before and after unrolling. Operations in the loop body become duplicated.}~\label{fig:unroll}
\end{figure}

This reduced the number of launched kernels by 10x and we saw a 3.5x speedup over the previously optimized but not unrolled implementation. While the downside of unrolling is increased program size (as seen in Figure \ref{fig:unroll}) and increased compile time (from roughly 300ms to 1400ms), the upside is that memory locations can be precomputed by the compiler, jumps to the start of the loop are eliminated, and intermediate between values stay in thread-local registers. We used Nsight Compute to confirm that the arithmetic intensity increased by 10x (because values are loaded once and operated on 10 times), memory unit stalls dropped by 5x, and achieved FLOP/s increased by 3.5x. We also saw math unit stalls increase by 2x, but this is because we are doing a better job keeping the Float32 compute units busy.

\subsection{Comparison with Jax CPU implementation}
We also compared the throughput speed of our fastest implementation (unroll 10) using the XLA CPU backend on an AMD Ryzen 7 5800X 8-core CPU. We found that the CPU is faster than the GPU when there are 70 or less simulators running in parallel. However, because the GPU is capable of higher thread parallelism, it achieves higher throughput when running more than 70 parallel simulators.

\subsection{Comparison with PyTorch and TorchScript implementations}
Our PyTorch implementation of cartpole eagerly executes tiny kernels for individual operations, leading to dozens more kernels launched than our XLA baseline. As a result we measured a slowdown to 0.13x compared to our baseline. However, TorchScript includes a compiler that performs instruction fusion. So we implemented cartpole in TorchScript and verified that it emits a single CUDA kernel that is fully fused. We found a 1.97x speedup and believe this is on a similar order of magnitude of throughput as the fully fused Jax-XLA kernel. However, Jax is much more flexible because it can differentiate functions and can use more of python's native features like loops, without breaking compilation.

\subsection{Comparison with CUDA implementation}

\begin{figure}[h]
\centering
  \includegraphics[width=1\columnwidth]{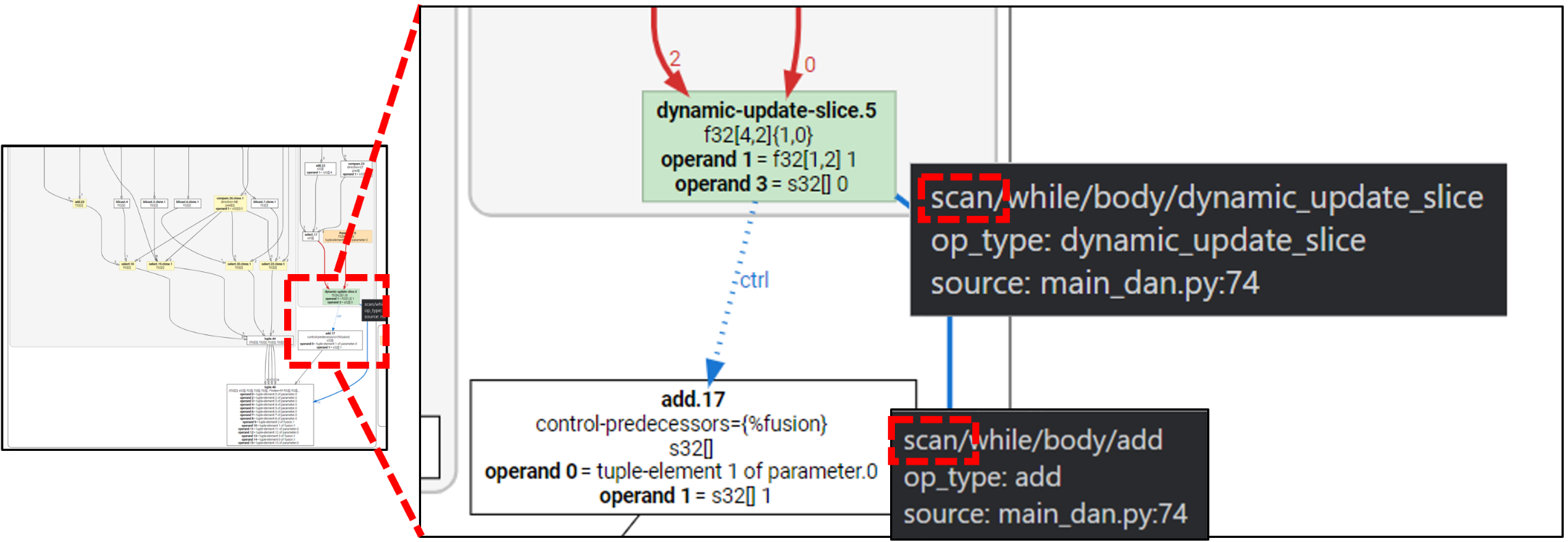}
  \caption{The source of XLA's main slowdown as compared to CUDA are these two extraneous kernels which are involved in XLA's implementation of loops. }~\label{fig:scan_kernels}
\end{figure}

An implementation of cartpole written in CUDA was contributed by James Gleeson. While we found that his implementation differs in the datatype that it used (his CUDA uses F16, F32, F64 and our Jax uses F32 datatype only), it is otherwise functionally equivalent to our Jax baseline. We find that the CUDA implementation is 2.7x faster than our best XLA implementation, and 28x faster than our naïve XLA implementation. We find that XLA introduces framework overhead and we used Nsight System to verify the speed differential is fully accounted for. Specifically, XLA launches extra 2 kernels (see Figure \ref{fig:scan_kernels}) per simulation loop iteration due to the way it implements lops. And these two kernels are launch overhead bound.

To provide more details, we found the runtime of the CUDA cartpole implementation to be 23\% slower for a single CUDA kernel to run 5 simulation steps, as compared to the Jax with 5 loop unrolls. However, it's a very different story for 10,000 steps. In that case, the CUDA implementation still runs only 1 kernel, while our best XLA implementation runs 3 kernels for each loop iteration, and requires 1000 iterations when we unroll by 10 to reach 10,000 steps, which totals 3,000 kernel launches for our best XLA implementation. These additional kernels are small but together add up to 2.7x longer runtime than the CUDA implementation.

%% file: discussion.tex
\section{Discussions}
In this section, we discuss the existing limitations of XLA compiler and potential directions for the future research.
\subsection{Limitations}
Though XLA provides a simplest way for ML developers and researchers to optimize their codes, XLA's generated code is still sub-optimal compared to other heavily tuned ML compilers. Combining with our investigations shown in the previous sections, we give the following limitations/challenges for XLA. \\
1) The performance of the generated kernel heavily relies on the quality of frontend Python code (e.g., concatenate operation in our Cartpole environment). This is not a unique problem for XLA, other ML compilers also have similar problems.
The fusion mechanism itself cannot address such frontend code quality issues, instead, it requires additional optimization passes that can diagnose inefficient frontend code and perform proper equivalent code transformation for better performance.\\
% here is another example for matrix multiplication XLA: https://zhuanlan.zhihu.com/p/88991966
2) The custom CUDA kernel calls limit the further fusion opportunities for XLA-GPU. Compared to XLA-TPU, XLA-GPU still uses third-party cuDNN/cuBLAS primitives for compute-intensive DL operations, which results in separate fused kernels and additional layout conversion overhead. Further optimization is obtainable if XLA has its own efficient implementations of tunable DL operations.\\
3) Rule-based fusion in XLA is inflexible and conservative. XLA's fusion rule needs to consider various cases that could appear in the HLO graph, thus XLA has relatively conservative fusion strategies that are guaranteed to not hurt the performance of the code. For example, the inability of XLA's fusion at the concatenate operation in our initial Cartpole implementation can be treated as overkill. The task-specific autotuning methods \cite{phothilimthana2021flexible,chen2018tvm,jung2021deepcuts} are able to address this problem and generate better fused kernels, but this can also result in much higher compilation overhead.\\
4) XLA relies on frontend code conversion and JIT compilation, which can introduce non-trivial runtime overhead for lightweight arithmetic computation with varied input shape. This is an inevitable tradeoff between flexibility and efficiency. For lightweight or varied programs, PyTorch-like eager execution may give better performance as compared to heavy JIT. 
\subsection{Future Work}
In addition to potential optimizations mentioned above, there are other future works we can explore:\\
1) Fusing simulation with neural network inference and extending our characterization of kernel fusion. DL models will make the study of fusion much more complex, and it would be interesting to see if there could be new optimization opportunities for ML compilers, for example, multi-stream parallelism for simulation and inference.\\
2) Auto-tuning the loop unrolling. We see a great performance gain from the loop unrolling. Due to the nature of iterative execution in ML tasks, automatic loop unrolling can be a good way to avoid frequent kernel stalls for repetitively executed kernels.\\
3) Fusion in DL training. Training support is one of the important advantages of XLA over other ML compilers. DL training requires the intermediate features for the backward gradient-based optimization, which limits the benefits of the operation fusion, and introduces other optimization options e.g., rematerialization. We have not done such characterization in this project, we will leave it as our future exploration.

% If we are successful we should find XLA's reasons for kernel fusion in cartpole, if they work as intended, if they are beneficial, and why XLA is slower than CUDA. We will also see if XLA can be modified to fuse cartpole into 1 kernel.

% We will present:
% \begin{enumerate}
% \item \textbf{A computational graph} of cartpole annotated with reasons why XLA was split into several kernels.
% \item \textbf{Performance graphs} showing key metrics that changed when we use XLA to fuse cartpole into 1 kernel. These graphs may include instructions per cycle (IPC), stall reasons, block/thread counts, and utilization of compute units, registers, and memory hierarchy. The final graphs depend on our findings.
% \end{enumerate}

% Actions to be taken in more detail:
% \begin{enumerate}
%   \item Determine why XLA cartpole is not fully fused.
%   \begin{enumerate}
%     \item Study verbose log output to uncover fusion decisions.
%   \end{enumerate}
%   \item Determine why XLA cartpole is not fully fused.
%   \begin{enumerate}
%     \item Modify the XLA source code or remove some XLA passes.
%     \item Profile with Nsight Compute to verify if XLA's fusion decisions actually had the effect it intended to have on low level hardware constraints.
%   \end{enumerate}
%   \item Determine where time is spent to see why XLA cartpole slower than CUDA cartpole.
%   \begin{enumerate}
%     \item Compare original XLA to our aggressive fusion modification.
%     \item Use Nsight System to record timings and stack traces or NVTX annotations to determine the purpose of the time spent.
%   \end{enumerate}
% \end{enumerate}

%% file: conclusion.tex
\section{Conclusions}
Throughout this paper we shed light on the concepts and inner workings of XLA's a poorly documented, but important optimization. Automatic kernel fusion is far from a solved problem. There are many fusion techniques (ie. instruction fusion,  fusion merger, sibling fusion, and producer-consumer fusion), who's automatic application is hindered by practical considerations like code duplication and hardware limits (eg. register size), and engineering limitations such as arbitrary expensive ops (eg. log, power) and in-fusable custom kernels (eg. cuRAND, cuDNN). We fused kernels in three different ways (modifying XLA, optimizing our code, and loop unrolling). Our key takeaway is that we were able to utilize kernel fusion in XLA for a speedup up to 10.5x, but XLA's has framework overhead that makes a handwritten CUDA implementation even faster.
\par
We hope that our insights into how XLA operates helps ML developers better understand the systems they use, and leads improvements in ML compilers, an open area of systems research. 